\documentclass{article}
\usepackage{spconf,amsmath,graphicx}
\usepackage{bm}
\usepackage{mathrsfs}
\usepackage{algorithm}
\usepackage{amsfonts} 
\usepackage{algorithmicx}
\usepackage{algpseudocode}
\usepackage{graphicx}
\usepackage{textcomp}
\usepackage{xcolor}
\usepackage{multirow}
\usepackage{booktabs}
\usepackage{caption2}
\usepackage{subfigure} 
\usepackage{threeparttable} 
\usepackage{amsthm}

\title{MHSCNet: A Multimodal Hierarchical Shot-aware Convolutional Network for Video Summarization}
\name{\begin{tabular}{c}Wujiang~Xu $^{\dagger}$,
        Runzhong~Wang $^{\star}$ ,
        Xiaobo~Guo $^\ddag$ $\S$ \thanks{\S\quad Corresponding author.},
        Shaoshuai~Li $^{\dagger}$,        
        Qiongxu~Ma $^{\dagger}$ \\
			Yunan~Zhao $^{\dagger}$,
        Sheng~Guo $^{\dagger}$,
        Zhenfeng Zhu $^{\ddag}$,
        Junchi~Yan $^{\star}$\end{tabular}}
\address{$^{\dagger}$	MYbank, Ant Group, Hangzhou, China\\
$^\ddag$ Institute of Information Science, Beijing Jiaotong University, Beijing, China \\
$^{\star}$	Department of Computer Science and Engineering, Shanghai Jiao Tong University, Shanghai, China}

\begin{document}
%
\maketitle
\begin{abstract}
Video summarization is an essential problem in signal processing, which intends to produce a concise summary of the original video. Existing video summarization approaches regard the task as a keyframe selection problem and generally construct the frame-wise representation by combining the long-range temporal dependency with either unimodal or bimodal information.
The optimal keyframe should offer the semantic summarization of the whole content by exploiting the multimodal and shot-level hierarchical natures of videos, however, such natures are not fully exploited in existing methods.
In this paper, we propose to construct a more powerful and robust frame-wise representation and predict the frame-level importance score in a fair and comprehensive manner.
Specifically, we propose a \underline{\textbf{m}}ultimodal \underline{\textbf{h}}ierarchical \underline{\textbf{s}}hot-aware \underline{\textbf{c}}onvolutional \underline{\textbf{net}}work, denoted as MHSCNet, to enhance the frame-wise representation via combining the comprehensive available multimodal information.
We further design a hierarchical ShotConv network to incorporate the adaptive shot-aware frame-level representation by considering the short-range and long-range temporal dependencies.
Based on the learned shot-aware representations, MHSCNet can predict the frame-level importance score in the local and global view of the video.
Extensive experiments on two standard video summarization datasets demonstrate that our proposed method consistently outperforms state-of-the-arts. 
\end{abstract}
\section{Introduction}
\label{sec:intro}
Recently, video summarization has gained increasing interests to figure out how to browse, manage and retrieve videos efficiently, which aims to create a short synopsis that preserves the most important and relevant content of the original video with minimal redundancy \cite{rochan2018video}. Typically, the most recent video summarizers follow three steps: 1) video segmentation, 2) importance score prediction, and 3) key shot selection. Regarding video segmentation and key shot selection, most of the previous methods \cite{park2020sumgraph} utilize KTS (Kernel Temporal Segmentation) and 0-1 knapsack algorithm 
to generate the summary results. The most challenging
step is importance score prediction, which is designed to highlight the most important parts according to the content.

Early efforts exploit LSTM units~\cite{zhang2016video,mahasseni2017unsupervised} to capture long-range temporal dependency among video frames. Some methods~\cite{he2019unsupervised,fajtl2018summarizing} resort to the attention mechanism to propagate the temporal dependency information in the frame level. Being aware of the inherent shot-level structure of videos, some researchers~\cite{liu2019learning,li2021exploring,jung2020global} investigate the short-term dependency and extract the shot-level representation via the attention mechanism. Since video temporal structures are intrinsically hierarchical, i.e., a video is composed of shots, and a shot is composed of several frames, the shot-level representation is beneficial for a video summary. However, the fixed number of the frames within a shot and the failure to classify between global and local information leads to the drawback of existing methods~\cite{liu2019learning,li2021exploring,jung2020global}, whereby the frame-level importance score cannot be achieved in a fair and global view, especially for videos with various lengths. Moreover, the above methods only adopt visual cues to construct the frame-wise representation, leading to insufficient frame-level representation. To resolve this issue, some bimodal methods \cite{wei2018video,zhao2021audiovisual,narasimhan2021clip} propose to fuse the visual modality with the language modality or the audio modality to enhance the frame representation. Since shot-level representation and multimodal information are two orthogonal efforts toward better video summarization, this paper combines the best of two worlds.

\begin{figure*}[h!]
\centering{
\includegraphics[width=0.95\textwidth]{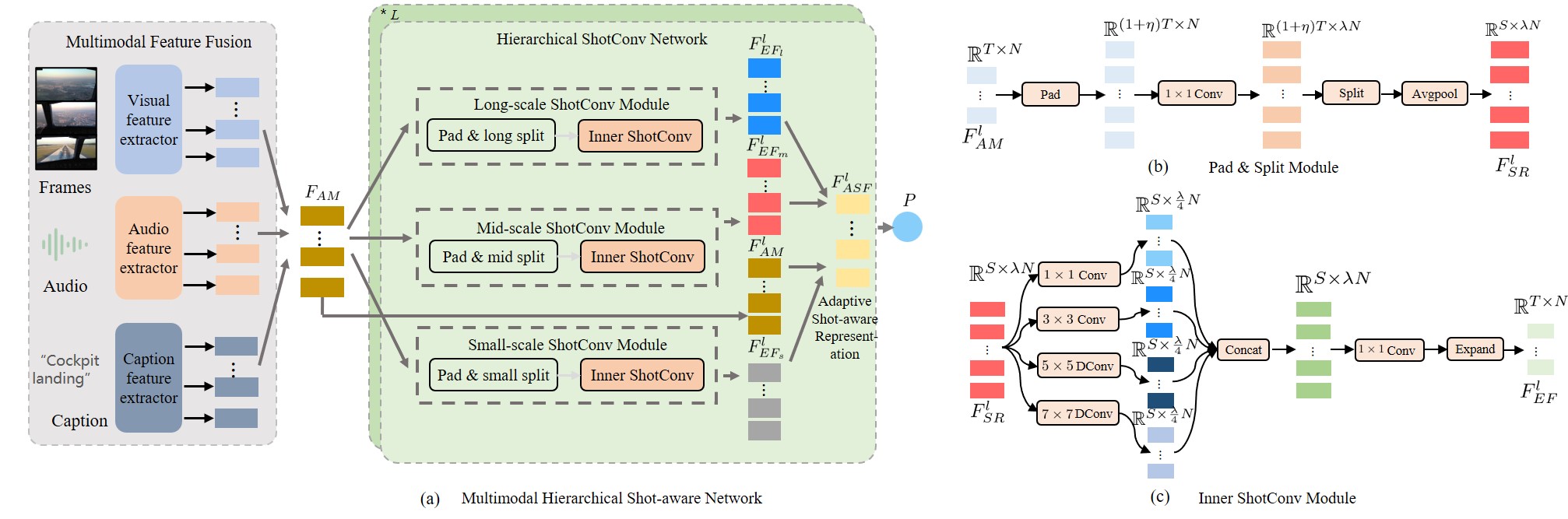}}
\vspace{-10pt}
\caption{(a) The overview of our multimodal hierarchical shot-aware convolutional network (MHSCNet). (b) The illustration of our cross-shot padding \& split module. (c) The illustration of our inner ShotConv module.}   
\label{Fig_framework}
\end{figure*}

To address the above challenges, we present a \underline{\textbf{m}}ultimodal \underline{\textbf{h}}ierarchical \underline{\textbf{s}}hot-aware \underline{\textbf{c}}onvolutional \underline{\textbf{net}}work, referring to \textbf{MHSCNet}. The shot-aware network means that the inner operation among the network is defined at the shot level. As illustrated in Fig.~\ref{Fig_framework}, it first encodes the multimodal inputs individually with pre-trained models to generate separate representations and then aggregates them through the designed cross-modal attention mechanism to explore the semantic relations.
MHSCNet fully exploits the cross-modality information to construct powerful frame-wise representation, by considering the audio and the caption of the video, in complement of the basic modality i.e.\ the video frames. Then, our hierarchical shot-aware convolutional network with the cross-shot padding mechanism provides an adaptive shot-aware representation with both short-range and long-range temporal dependency. Different from previous methods \cite{apostolidis2021combining,liu2019learning,fu2021video}, our hierarchical architecture can generate the summary results better for videos with various lengths.

The contributions of this work are summarized as follows: 

- We propose a multimodal framework which utilizes all available modalities with a cross-modal attention mechanism to fully explore the semantic relations among the modalities.

- We design a hierarchical ShotConv network with the cross-shot padding mechanism to produce the adaptive shot-aware representation. Specifically, our ShotConv differs from \cite{apostolidis2021combining,liu2019learning,fu2021video}: 1) The ShotConv network is a hierarchical design which made the shot-level representation be adaptive to videos with various lengths; 2) The cross-shot padding mechanism helps the shot-level representation containing the global information and the local information together after the stacked ShotConv network.

- The experiment results indicate that our MHSCNet outperforms existing SOTA methods on two open benchmarks. Our code is available at the link \footnote{https://github.com/WujiangXu/MHSCNet}.

\section{Related Work}
Early efforts of video summarization \cite{zhang2016video,fajtl2018summarizing} exploit the attention mechanism or the LSTM units to capture long-range temporal dependency. To better model the semantic relation among the frames, Park et al.~\cite{park2020sumgraph} propose a recursive graph modeling networks for video summarization. However, only adopting the long-range temporal dependency overlooks the frame-frame relation within the shot leading to insufficient frame-wise representation. Therefore, some recent works  \cite{liu2019learning,li2021exploring,jung2020global} start to design the model to extract the shot-level representation. Liu et al.~\cite{liu2019learning} utilize the multi-head attention mechanism to learn the shot-level representation. Similarly, Jung et al.~\cite{jung2020global} design a relative position embedding to enhance the self-attention network performance. Though these works show the importance of exploring the short-range temporal dependency within the shot, they fix the length of the shot, causing a worse performance for different videos with various lengths. Moreover, the above methods only use the visual modality to construct the frame-wise representation, which ignores the other modalities of the video. 
To build powerful frame-wise representation, some researchers  \cite{zhou2018video,wei2018video,zhao2021audiovisual} investigate the power of the language or audio modalities. Recently, Narasimhan et al.~\cite{narasimhan2021clip} propose a language-guided transformer for video summarization.
Nevertheless, they do not fully utilize the modality information of the video. Our MHSCNet fully exploits the modality information including audio and caption of the video via the designed cross-modal attention mechanism.
 
\section{Methodology}
\label{sec:method}

\subsection{Multimodal Feature Fusion}
Given a video $V$ as a sequence of $T$ frames, we first employ CNN (e.g. GoogLeNet) to extract feature $F_{V} =[f_{v_{1}}, ..., f_{v_{i}}, ..., f_{v_{T}}] \in \mathbb{R}^{T\times N}$. For audio information  $F_{A} =[f_{a_{1}}, ..., f_{a_{i}},$ $..., f_{a_{T}}] \in \mathbb{R}^{T\times M}$, the VGGish \cite{hershey2017cnn} is adopted for audio feature extraction. To capture generic caption information, we use the Bi-Modal Transformer \cite{iashin2020better} to generate video captions. We use the CLIP \cite{radford2021learning} model to encode caption vector $F_{C} \in \mathbb{R}^{1 \times N}$ and average pooling is used to balance multiple sentences. 
Then, we fuse the image feature with the audio feature. For the $i$-th frame, we project the audio vector $f_{a_{i}}$ to the same embedding space with the image feature $F_{V}$ by a trainable matrix $W_{audio}$ like 
$f_{va_{i}} = f_{v_{i}} + f_{a_{i}}W_{audio} +b_{audio}$,
where $W_{audio} \in \mathbb{R}^{M \times N}$ is the trainable matrix and $b_{audio}\in \mathbb{R}^{N}$ is the trainable bias.
After fusing the audio information, we design a cross-modal attention mechanism. We regard the caption feature $F_{C}$ as both the Key and the Value and the enhanced local feature $F_{VA}$ as the Query. Therefore, the aggregated multimodal frame-level feature $F_{AM}$ can be obtained through the multi-head attention mechanism.

\subsection{Hierarchical ShotConv Network}
\noindent \textbf{Cross-shot Padding \& Split.} \
As shown in Fig.~\ref{Fig_framework}(b), we firstly pad the specified number of frames in the boundary of each shot. Given a padding ratio $\eta$ and the shot number $S$, $\eta \times \lfloor \frac{T}{S} \rfloor$ frames from the end of the $(S$-1)-th shot will be padded at the start of the $S$-th shot. Besides, we also pad the parts of frames from the last shot to the start of the first shot. Then a 1$\times$1 convolution is utilized to increase the channel of the frame from $N$ to $\lambda  N$ and the frame-level feature $F_{CP}^{l} =[f_{cp_{1}}^{l},...,f_{cp_{i}}^{l},...,f_{cp_{(1+\eta) T}}^{l}] \in \mathbb{R}^{(1+\eta) T\times \lambda  N}$ in the $l$-th layer could be obtained via the cross-shot padding mechanism. Then, we reshape the dimension from $(1+\eta) T\times \lambda  N$ to $S \times (D_{S} \cdot \lambda  N)$, where $D_{S} = \lfloor \frac{(1+\eta) T}{S} \rfloor$ and $\lfloor . \rfloor$ is the round-down function to transform the frame-level representation to the inflated shot-level representation $F_{ISR}^{l}$. Specially, the $s$-th shot contains the features of inside frames as follows:
\begin{equation}
F_{ISR_{s}}^{l} = \{f_{cp_{k}}^{l}: k=(s-1)D_{S} + 1 ,...,sD_{S} ,sD_{S} +1   \} 
\label{shot_split}
\end{equation}
Lastly, an average pooling operator will be used on the dimension $D_{S}$ of the frame number to obtain the shot-level representation $F_{SR}^{l} \in \mathbb{R}^{S \times \lambda  N}$. To make our network adaptive to the video length, we propose hierarchical cross-shot padding, and the shot number $S$ will be set to $S_{l}$, $S_{m}$, $S_{s}$ for the long scale, middle scale, and short scale to generate the corresponding representation $F_{SR_{l}}^{l}$, $F_{SR_{m}}^{l}$ and $F_{SR_{s}}^{l}$. 

\noindent \textbf{Inner ShotConv Module.} \
The inner shot convolution module is composed of four convolution operators with different kernel sizes and other operations. The strides of all convolution operators are set to 4. The 5$\times$5 and 7$\times$7 convolution operators are implemented by the dilated convolution \cite{yu2015multi}. We first perform these four convolution operators on $F_{SR}^{l}$. 
Then we will concatenate them in the channel dimension and feed it to a 1$\times$1 convolution layer to reduce the channel dimension from $\lambda N$ to $N$. Each shot-level representation is assigned to the frames in the same shot to get the enhanced frame-level representation $F_{EF}^{l}$ via the expand operation. For the three scales ShotConv modules, the enhanced frame-level representation $F_{EF_{l}}^{l}$, $F_{EF_{m}}^{l}$ and $F_{EF_{s}}^{l}$ are generated. The adaptive shot-aware representation $F_{ASF}^{l} \in \mathbb{R}^{T \times N}$ is obtained via $F_{ASF}^{l} = F_{EF_{l}}^{l} + F_{EF_{m}}^{l} + F_{EF_{s}}^{l} + F_{AM}^{l}$, where $F_{AM}^{l}$ is replaced by $F_{ASF}^{l-1}$ except the first layer. We stack $L$ layers of the hierarchical ShotConv network and the output from each layer is the input of the next layer.

\noindent \textbf{Implicit Global Information Propagation.} \
In each layer of the ShotConv network, the frames in the local shot learn the information from the other adjacent shot via the proposed cross-shot padding mechanism. The frames in the local shot learns the global information from the whole video by stacked ShotConv networks. Formally, the global information is passed in a local shot when the number of stacked layers and the shot number meet the criteria $L+1 >= S_{l}$, where $L$ is the number of the stacked hierarchical ShotConv network and $S_{l}$ is the shot number of the long scale ShotConv module.

\subsection{End-to-End Supervised Learning}
All modules in our pipeline support gradient back propagation. We apply the focal loss \cite{lin2017focal} for keyframe classification, which resolves the imbalance between the number of keyframes and the number of background frames.

\section{Experiments}
\subsection{Dataset}
Following previous works \cite{narasimhan2021clip,park2020sumgraph}, We evaluate the performance on two datasets: SumMe \cite{gygli2014creating} and TVSum \cite{song2015tvsum} under the three different data configuration setting including Standard, Augment and Transfer data setting \cite{zhang2016summary,rochan2018video}.

\subsection{Experiment Setting}
We use the ImageNet-pretrained GoogleNet \cite{szegedy2015going} for frame feature extraction, where the 1024-d feature are extracted after the `pool5' layer. The dimension of the audio feature extracted by the pretrained VGGish is 128. For the generated multiple captions by the BMT \cite{iashin2020better}, we use CLIP \cite{radford2021learning} to encode them and concatenate the embedding in the channel dimension. The dimension of the first fully-connected layer is 128. The number of multi-head in the caption feature fusion module is set to 32.
The different time scales for defining the shots follow the relationship $S_{s} = 1.5 S_{m} =3 S_{l}$. We tune $S_{l}$ on dataset SumMe. We set $ S_{l} = 5$, $S_{m}=10$, $S_{s}=15$, $\eta=0.05$, $L = 4$ and $\lambda = 8$.
$\alpha$ and $\gamma$ in the focal loss are set to 0.25 and 2. All the experiment results are reported in 5-fold cross-validation. Adam optimizer is adopted with an initial learning rate of $10^{-4}$. We employ F-score as the evaluation metric for the quality of the generated summary.

\subsection{Comparisons to the State-of-the-Arts}
\begin{table*}[htbp!]
\footnotesize
\caption{Comparisons of $F$-Score (\%) and parameters (in millions) with other methods on SumMe \cite{gygli2014creating} and TVSum \cite{song2015tvsum}. Bimodal approaches are marked with $\diamond$. As CLIP-It, we report their results with two different backbone networks. The missing results are due to no open source code.}
    \centering
    \setlength\tabcolsep{9pt}{
\begin{tabular}{lcccccccc}
\toprule[0.75pt]
\multicolumn{1}{l}{\multirow{2}{*}{Method}} &\multicolumn{1}{c}{\multirow{2}{*}{Backbone}} & \multicolumn{3}{c}{SumMe}                                                                   & \multicolumn{3}{c}{TVSum} & \multicolumn{1}{c}{\multirow{2}{*}{Params}}                                                                   \\ \cline{3-8} 
\multicolumn{1}{l}{}        & \multicolumn{1}{l}{}                & \multicolumn{1}{l}{Standard} & \multicolumn{1}{l}{Augment} & \multicolumn{1}{l}{Transfer} & \multicolumn{1}{l}{Standard} & \multicolumn{1}{l}{Augment} & \multicolumn{1}{l}{Transfer} \\ \hline\hline
SUM-GAN \cite{mahasseni2017unsupervised}     & GoogLeNet \cite{szegedy2015going}                                   & 41.7                          & 43.6                         & -                             & 56.3                          & 61.2                         & -                &295.9             \\
SumGraph \cite{park2020sumgraph}       &GoogLeNet \cite{szegedy2015going}                                & 51.4                          & 52.9                         & 48.7                          & 63.9                          & 65.8                         & 60.5                & 5.5         \\
H-MAN \cite{liu2019learning}       &GoogLeNet \cite{szegedy2015going}                       & 51.8                          & 52.5                         & 48.1                          & 60.4                          & 61.0                         & 59.5                & -         \\
CSNet+GL-RPE \cite{jung2020global}   &GoogLeNet \cite{szegedy2015going}      &50.2                    &-                          & -                        & 59.1                          & -                         & -                         & -                      \\
SUM-GDA \cite{li2021exploring}   &GoogLeNet \cite{szegedy2015going}      &52.8                     & 54.4                          & 46.9                                                 & 58.9                          & 60.1        & 59.0                & -                      \\
 \hline
$\diamond$DQSN \cite{zhou2018video}   &GoogLeNet \cite{szegedy2015going}                                  & -                          & -                         & -                          & 58.6                         & -                         & -                & -          \\
$\diamond$SASUM \cite{wei2018video}    &GoogLeNet \cite{szegedy2015going}                                 & 45.3                         & -                         & -                          & 58.2                        & -                         & -                & -          \\
$\diamond$AVRN \cite{zhao2021audiovisual}    &GoogLeNet \cite{szegedy2015going}                                  & 44.1                          & 44.9                        & 43.2                          & 59.7                     & 60.5                         & 58.7                & -          \\
$\diamond {\rm CLIP-It}^{*}$ \cite{narasimhan2021clip}     &GoogLeNet  \cite{szegedy2015going}                                  & 51.6                          & 53.5                         & 49.4                         & 64.2                          & 66.3                        & 61.3                & -         \\
$\diamond {\rm CLIP-It}$ \cite{narasimhan2021clip}      &CLIP-ViT-B/32 \cite{radford2021learning}                                & 54.2                          & 56.4                         & \textbf{51.9}                         & 66.3                          & 69.0                        & \textbf{65.5}                & -         \\\hline
MHSCNet (ours)    &GoogLeNet \cite{szegedy2015going}                                        & \textbf{55.3}                          &\textbf{56.9}                              & 49.8                              & \textbf{69.3}                         &  \textbf{69.4}                              & 61.0     &134.6           \\ \toprule[0.750pt]
\end{tabular}}
\vspace{-10pt}
\label{table1}
\end{table*}
\noindent \textbf{Quantitative comparison.} \ 
As reported in Table \ref{table1}, our MHSCNet achieves a significant improvement ranging from 1.1\% to 13.6\% compared to the SOTA methods in the standard data setting equipped with the same backbone network. The compared methods can be roughly classified into two categories: 1) unimodal methods (SUM-GAN \cite{mahasseni2017unsupervised}, SumGraph  \cite{park2020sumgraph}, H-MAN  \cite{liu2019learning} , CSNet+GL-RPE \cite{jung2020global}, SUM-GDA  \cite{li2021exploring}), 2) bimodal methods (DQSN  \cite{zhou2018video}, SASUM \cite{wei2018video}, AVRN  \cite{zhao2021audiovisual} , CLIP-It \cite{narasimhan2021clip}). By exploring the relation within the shot, H-MAN  \cite{liu2019learning} , CSNet+GL-RPE \cite{jung2020global} and SUM-GDA  \cite{li2021exploring} achieve better performance than SUM-GAN \cite{mahasseni2017unsupervised} and SumGraph \cite{park2020sumgraph} in most cases, which indicates that shot-level representation is important for improving the prediction accuracy. 
Without modeling the relationship between the frames well, these three bimodal methods (DQSN  \cite{zhou2018video}, SASUM \cite{wei2018video} and AVRN  \cite{zhao2021audiovisual}) are even inferior than the unimodal methods. Benefitting from the language-guided transformers and the self-attention mechanism, CLIP-It achieves the second-best performance. However, they ignore the audio modality information and the frame-frame relation within the shot leading to a worse performance than our MHSCNet.
By constructing a hierarchical shot-level representation and fully exploiting the modality information of the video by our cross-modal attention mechanism, our model MHSCNet achieves the best performance equipped with the same backbone network.

\subsection{Ablation Study}
We conduct a series of experiments on the SumMe  \cite{gygli2014creating} dataset and the TVSum \cite{song2015tvsum} datasets to better understand the proposed model and verify the contribution of each component. We first study the influence of the additional modality information of our MHSCNet. Without the audio modality and the caption modality, our model also achieve the best results on the TVSum dataset compared to the unimodal methods. Besides, we study the influence at different scales of the shots in the hierarchical ShotConv network. Compared to the double-scale ShotConv networks, the single-scale ShotConv networks perform worse. However, the network with the single middle-scale shot convolution still achieves competitive performance due to the split shot length being matched with the initial shot length of the video. Similarly, the network with a single long-scale shot convolution obtains a 67.6\% F-score which is slightly lower than the best results of the double-scale networks. Moreover, we observe that a double-scale network such as M+S drops the performance compared to a single-scale network such as M. The negative transfer occasionally happens in the double-scale network. In brief, our MHSCNet equipped with all modality information and three scales ShotConv modules achieves the best results.
\begin{table}[h!]
\footnotesize
\caption{Ablation study in our MHSCNet with F-score metric on SumMe \cite{gygli2014creating} and TVSum \cite{song2015tvsum}.}
    \centering
    \setlength\tabcolsep{7pt}{
    \begin{tabular}{lcc}
    \toprule[0.75pt]
        Model & SumMe & TVSum\\ \hline\hline
        Modality &  &  \\
        Image  & 52.0  & 68.0  \\ 
        Image+Audio  & 54.5 & 68.2  \\ 
        Image+Caption  & 53.8 & 68.3  \\   \hline
        Scale &  &   \\
        Long-scale ShotConv module  & 52.2  & 67.6  \\ 
        Middle-scale ShotConv module  & 54.1 & 67.2  \\ 
        Short-scale ShotConv module  & 52.1 & 65.4  \\ 
        Long-scale+Middle-scale ShotConv modules  &54.4  & 67.5  \\ 
        Long-scale+Short-scale ShotConv modules  &53.0  & 66.0  \\ 
        Middle-scale+Short-scale ShotConv modules  & 53.6 & 67.7  \\   \hline
        Ours (Image+Audio+Caption) & \textbf{55.3} & \textbf{69.3}  \\ \toprule[0.75pt]
    \end{tabular}}
    \vspace{-10pt}
    \label{abl_modal}
\end{table}

\section{Conclusion}
In this paper, we propose a novel multimodal framework with a hierarchical ShotConv network for video summarization. Unlike existing unimodal or bimodal methods which only adopt visual cues with one additional modal information to construct the frame representation, our MHSCNet exploits the caption and the audio modality to enhance the frame-wise representation and explore the semantic relations via the proposed cross-modal attention mechanism. To evaluate the frame-level importance score in a fair and comprehensive manner, we further propose a hierarchical ShotConv network with the cross-shot padding mechanism to model the frame-to-frame interaction both within and across the shot. The hierarchical ShotConv network generates an adaptive shot-aware representation containing both short-range temporal dependency and long-range temporal dependency. 

\vfill\pagebreak

\bibliographystyle{IEEEtran}
\bibliography{IEEEabrv}

\begin{thebibliography}{10}
\providecommand{\url}[1]{#1}
\csname url@samestyle\endcsname
\providecommand{\newblock}{\relax}
\providecommand{\bibinfo}[2]{#2}
\providecommand{\BIBentrySTDinterwordspacing}{\spaceskip=0pt\relax}
\providecommand{\BIBentryALTinterwordstretchfactor}{4}
\providecommand{\BIBentryALTinterwordspacing}{\spaceskip=\fontdimen2\font plus
\BIBentryALTinterwordstretchfactor\fontdimen3\font minus
  \fontdimen4\font\relax}
\providecommand{\BIBforeignlanguage}[2]{{%
\expandafter\ifx\csname l@#1\endcsname\relax
\typeout{** WARNING: IEEEtran.bst: No hyphenation pattern has been}%
\typeout{** loaded for the language `#1'. Using the pattern for}%
\typeout{** the default language instead.}%
\else
\language=\csname l@#1\endcsname
\fi
#2}}
\providecommand{\BIBdecl}{\relax}
\BIBdecl

\bibitem{rochan2018video}
M.~Rochan, L.~Ye, and Y.~Wang, ``Video summarization using fully convolutional
  sequence networks,'' in \emph{ECCV}, 2018, pp. 347--363.

\bibitem{park2020sumgraph}
J.~Park, J.~Lee, I.-J. Kim, and K.~Sohn, ``Sumgraph: Video summarization via
  recursive graph modeling,'' in \emph{ECCV}.\hskip 1em plus 0.5em minus
  0.4em\relax Springer, 2020, pp. 647--663.

\bibitem{zhang2016video}
K.~Zhang, W.-L. Chao, F.~Sha, and K.~Grauman, ``Video summarization with long
  short-term memory,'' in \emph{European conference on computer vision}.\hskip
  1em plus 0.5em minus 0.4em\relax Springer, 2016, pp. 766--782.

\bibitem{mahasseni2017unsupervised}
B.~Mahasseni, M.~Lam, and S.~Todorovic, ``Unsupervised video summarization with
  adversarial lstm networks,'' in \emph{Proceedings of the IEEE conference on
  Computer Vision and Pattern Recognition}, 2017, pp. 202--211.

\bibitem{he2019unsupervised}
X.~He, Y.~Hua, T.~Song, Z.~Zhang, Z.~Xue, R.~Ma, N.~Robertson, and H.~Guan,
  ``Unsupervised video summarization with attentive conditional generative
  adversarial networks,'' in \emph{Proceedings of the 27th ACM International
  Conference on Multimedia}, 2019, pp. 2296--2304.

\bibitem{fajtl2018summarizing}
J.~Fajtl, H.~S. Sokeh, V.~Argyriou, D.~Monekosso, and P.~Remagnino,
  ``Summarizing videos with attention,'' in \emph{Asian Conference on Computer
  Vision}.\hskip 1em plus 0.5em minus 0.4em\relax Springer, 2018, pp. 39--54.

\bibitem{liu2019learning}
Y.-T. Liu, Y.-J. Li, F.-E. Yang, S.-F. Chen, and Y.-C.~F. Wang, ``Learning
  hierarchical self-attention for video summarization,'' in \emph{2019 IEEE
  international conference on image processing (ICIP)}.\hskip 1em plus 0.5em
  minus 0.4em\relax IEEE, 2019, pp. 3377--3381.

\bibitem{li2021exploring}
P.~Li, Q.~Ye, L.~Zhang, L.~Yuan, X.~Xu, and L.~Shao, ``Exploring global diverse
  attention via pairwise temporal relation for video summarization,''
  \emph{Pattern Recognition}, vol. 111, p. 107677, 2021.

\bibitem{jung2020global}
Y.~Jung, D.~Cho, S.~Woo, and I.~S. Kweon, ``Global-and-local relative position
  embedding for unsupervised video summarization,'' in \emph{European
  Conference on Computer Vision}.\hskip 1em plus 0.5em minus 0.4em\relax
  Springer, 2020, pp. 167--183.

\bibitem{wei2018video}
H.~Wei, B.~Ni, Y.~Yan, H.~Yu, X.~Yang, and C.~Yao, ``Video summarization via
  semantic attended networks,'' in \emph{Proceedings of the AAAI conference on
  artificial intelligence}, vol.~32, no.~1, 2018.

\bibitem{zhao2021audiovisual}
B.~Zhao, M.~Gong, and X.~Li, ``Audiovisual video summarization,'' \emph{arXiv
  preprint arXiv:2105.07667}, 2021.

\bibitem{narasimhan2021clip}
M.~e.~a. Narasimhan, ``Clip-it! language-guided video summarization,'' in
  \emph{NeurIPS}, 2021.

\bibitem{apostolidis2021combining}
E.~Apostolidis, G.~Balaouras, V.~Mezaris, and I.~Patras, ``Combining global and
  local attention with positional encoding for video summarization,'' in
  \emph{2021 IEEE ISM}.\hskip 1em plus 0.5em minus 0.4em\relax IEEE, 2021, pp.
  226--234.

\bibitem{fu2021video}
H.~Fu, H.~Wang, and J.~Yang, ``Video summarization with a dual attention
  capsule network,'' in \emph{2020 25th International Conference on Pattern
  Recognition (ICPR)}.\hskip 1em plus 0.5em minus 0.4em\relax IEEE, 2021, pp.
  446--451.

\bibitem{zhou2018video}
K.~Zhou, T.~Xiang, and A.~Cavallaro, ``Video summarisation by classification
  with deep reinforcement learning,'' \emph{arXiv preprint arXiv:1807.03089},
  2018.

\bibitem{hershey2017cnn}
S.~Hershey, S.~Chaudhuri, D.~P. Ellis, J.~F. Gemmeke, A.~Jansen, R.~C. Moore,
  M.~Plakal, D.~Platt, R.~A. Saurous, B.~Seybold \emph{et~al.}, ``Cnn
  architectures for large-scale audio classification,'' in \emph{2017 ieee
  international conference on acoustics, speech and signal processing
  (icassp)}.\hskip 1em plus 0.5em minus 0.4em\relax IEEE, 2017, pp. 131--135.

\bibitem{iashin2020better}
V.~Iashin and E.~Rahtu, ``A better use of audio-visual cues: Dense video
  captioning with bi-modal transformer,'' \emph{arXiv preprint
  arXiv:2005.08271}, 2020.

\bibitem{radford2021learning}
A.~Radford, J.~W. Kim, C.~Hallacy, A.~Ramesh, G.~Goh, S.~Agarwal, G.~Sastry,
  A.~Askell, P.~Mishkin, J.~Clark \emph{et~al.}, ``Learning transferable visual
  models from natural language supervision,'' in \emph{International Conference
  on Machine Learning}.\hskip 1em plus 0.5em minus 0.4em\relax PMLR, 2021, pp.
  8748--8763.

\bibitem{yu2015multi}
F.~Yu and V.~Koltun, ``Multi-scale context aggregation by dilated
  convolutions,'' \emph{arXiv preprint arXiv:1511.07122}, 2015.

\bibitem{lin2017focal}
T.-Y. Lin, P.~Goyal, R.~Girshick, K.~He, and P.~Doll{\'a}r, ``Focal loss for
  dense object detection,'' in \emph{Proceedings of the IEEE international
  conference on computer vision}, 2017, pp. 2980--2988.

\bibitem{gygli2014creating}
M.~Gygli, H.~Grabner, H.~Riemenschneider, and L.~Van~Gool, ``Creating summaries
  from user videos,'' in \emph{European conference on computer vision}.\hskip
  1em plus 0.5em minus 0.4em\relax Springer, 2014, pp. 505--520.

\bibitem{song2015tvsum}
Y.~Song, J.~Vallmitjana, A.~Stent, and A.~Jaimes, ``Tvsum: Summarizing web
  videos using titles,'' in \emph{CVPR}, 2015, pp. 5179--5187.

\bibitem{zhang2016summary}
K.~Zhang, W.-L. Chao, F.~Sha, and K.~Grauman, ``Summary transfer:
  Exemplar-based subset selection for video summarization,'' in \emph{CVPR},
  2016, pp. 1059--1067.

\bibitem{szegedy2015going}
C.~Szegedy, W.~Liu, Y.~Jia, and et~al., ``Going deeper with convolutions,'' in
  \emph{CVPR}, 2015, pp. 1--9.

\end{thebibliography}

\end{document}